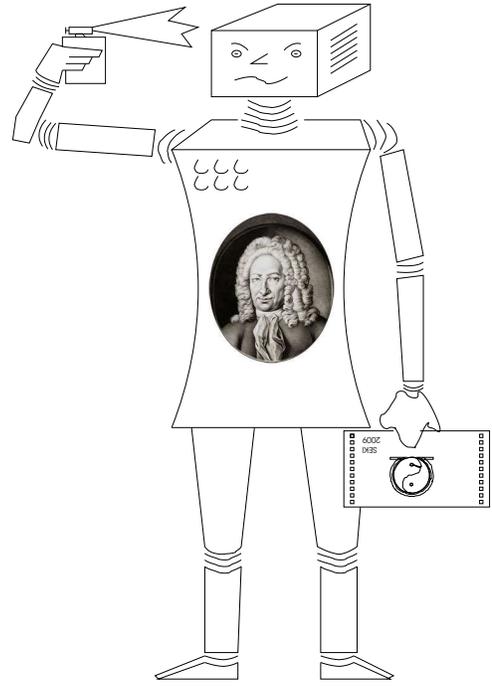

# Automating Quantified Multimodal Logics in Simple Type Theory

—

## A Case Study

Christoph Benzmüller

SEKI Working-Paper SWP–2009–02

# Automating
# Quantified Multimodal Logics
# in Simple Type Theory
# —
# A Case Study

Christoph Benzmüller

**Abstract**

In a case study we investigate whether off the shelf higher-order theorem provers and model generators can be employed to automate reasoning in and about quantified multimodal logics. In our experiments we exploit the new TPTP infrastructure for classical higher-order logic.

# 1  Introduction

This paper presents a case study in quantified multimodal logics. An interesting aspect of this case study is that off the shelf theorem provers and model generators for simple type theory, that is, classical higher-order logic, are employed to automate problems in quantified multimodal logics, that is, non-classical logics. This is enabled by our recent embedding of normal quantified multimodal logics in simple type theory [8, 10], which is sound and complete [10]. Interestingly, not only reasoning *within* various non-classical logics can be automated this way but also reasoning *about* them. For example, the equivalence between different properties of accessibility relations and their associated multimodal axioms can be proved automatically.

Generally, it seems that proof assistants and automated reasoning systems for classical higher-order logics can serve — ideally in combination — as a fruitful basis for the exploration and the modeling of normal multimodal logics, for the investigation of their meta-properties, and probably even for the application of these logics to real world problems. In fact, normal multimodal logics are simply fragments of classical higher-order logics with specific computational properties. Detecting and characterizing these fragments, in particular decidable ones, can be very beneficial for the higher-order reasoners themselves, since this may stimulate significant improvements of the reasoning procedures used within these reasoners.

The modeling of a specific propositional or quantified multimodal logic of interest in our framework is straightforward and not very time consuming. This is in contrast with the usual situation in the area of quantified and propositional multimodal logics, where a vast amount of specific calculi have been developed and published, while very few of them have actually been implemented and applied. The main reason is that the effective implementation of these calculi is usually non-trivial.

The paper is structured as follows: In Section 2 we present some preliminaries and our embedding of quantified multimodal logic into simple type theory. The new content, our small but interesting case study, is presented in Section 3 and an outlook is presented in Section 4.[1]

---

[1] Initially our plan was to extend the case presented here to propositional and quantified temporal logics. A study in (quantified) temporal logics will thus remain future work.)

## 2  $\mathcal{QML}$ as a fragment of $\mathcal{STT}$

$\mathcal{STT}$ [15] is based on the simply typed $\lambda$-calculus. The set $\mathcal{T}$ of simple types is usually freely generated from a set of basic types $\{o, \iota\}$ (where $o$ is the type of Booleans and $\iota$ is the type of individuals) using the function type constructor $\rightarrow$. Instead of $\{o, \iota\}$ we here consider a set of base types $\{o, \iota, \mu\}$, providing an additional base type $\mu$ (the type of possible worlds).

The simple type theory language $\mathcal{STT}$ is defined by ($\alpha, \beta \in \mathcal{T}$):

$$s,t ::= p_\alpha \mid X_\alpha \mid (\lambda X_\alpha . s_\beta)_{\alpha\to\beta} \mid (s_{\alpha\to\beta}\, t_\alpha)_\beta \mid (\neg_{o\to o}\, s_o)_o \mid$$
$$(s_o \vee_{o\to o\to o} t_o)_o \mid (s_\alpha =_{\alpha\to\alpha\to o} t_\alpha)_o \mid (\Pi_{(\alpha\to o)\to o}\, s_{\alpha\to o})_o$$

$p_\alpha$ denotes typed constants and $X_\alpha$ typed variables (distinct from $p_\alpha$). Complex typed terms are constructed via abstraction and application. Our logical connectives of choice are $\neg_{o\to o}$, $\vee_{o\to o\to o}$, $=_{\alpha\to\alpha\to o}$ and $\Pi_{(\alpha\to o)\to o}$ (for each type $\alpha$).[2] From these connectives, other logical connectives can be defined in the usual way (e.g., $\wedge$ and $\implies$). We often use binder notation $\forall X_\alpha . s$ for $\Pi_{(\alpha\to o)\to o}(\lambda X_\alpha . s_o)$. We assume familiarity with $\alpha$-conversion, $\beta$- and $\eta$-reduction, and the existence of $\beta$- and $\beta\eta$-normal forms. Moreover, we obey the usual definitions of free variable occurrences and substitutions.

The semantics of $\mathcal{STT}$ is well understood and thoroughly documented in the literature [2, 3, 7, 19]. The semantics of choice for our work is Henkin semantics.

$\mathcal{QML}$ has been studied by Fitting [16] (further related work is available by Blackburn and Marx [13] and Braüner [14]). In contrast to Fitting we are here not interested only in **S5** structures but in the more general case of **K** from which more constrained structures (such as **S5**) can be easily obtained. First order quantification can be constant domain or varying domain. Below we only consider the constant domain case: every possible world has the same domain. Like Fitting, we keep our definitions simple by not having function or constant symbols. While Fitting [16] studies quantified monomodal logic, we are interested in quantified multimodal logic. Hence, we introduce multiple $\square_r$ operators for symbols $r$ from an index set $S$. The grammar for our quantified multimodal logic $\mathcal{QML}$ hence is

$$s,t ::= P \mid k(X^1, \ldots, X^n) \mid \neg\, s \mid s \vee t \mid \forall X . s \mid \forall P . s \mid \square_r\, s$$

where $P$ denotes propositional variables, $X, X^i$ denote first-order (individual) variables, and $k$ denotes predicate symbols of any arity. Further connectives, quantifiers, and modal operators can be defined as usual. We also obey the usual definitions of free variable occurrences and substitutions.

Fitting introduces three different notions of Kripke semantics for $\mathcal{QML}$: **QS5**$\pi^-$, **QS5**$\pi$, and **QS5**$\pi^+$. In our work [10] we study related notions **QK**$\pi^-$, **QK**$\pi$, and **QK**$\pi^+$ for a modal context **K**, and we support multiple modalities.

$\mathcal{STT}$ is an expressive logic and it is thus not surprising that $\mathcal{QML}$ can be elegantly modeled and even automated as a fragment of $\mathcal{STT}$. The idea of the encoding, called $\mathcal{QML}^{STT}$, is simple. Choose type $\iota$ to denote the (non-empty) set of individuals and we reserve a second base type $\mu$ to denote the (non-empty) set of possible worlds. The type $o$ denotes the set of truth values. Certain formulas of type $\mu \to o$ then correspond to multimodal logic expressions. The multimodal connectives $\neg$, $\vee$, and $\square$, become $\lambda$-terms of types $(\mu \to o) \to (\mu \to o)$, $(\mu \to o) \to (\mu \to o) \to (\mu \to o)$, and $(\mu \to \mu \to o) \to (\mu \to o) \to (\mu \to o)$ respectively.

---

[2]This choice is not minimal (from $=_{\alpha\to\alpha\to o}$ all other logical constants can already be defined [4]). It useful though in the context of resolution based theorem proving.

Quantification is handled as in $\mathcal{STT}$ by modeling $\forall X_{\bullet} p$ as $\Pi(\lambda X_{\bullet}.p)$ for a suitably chosen connective $\Pi$. Here we are interested in defining two particular modal $\Pi$-connectives: $\mathbf{\Pi}^{\iota}$, for quantification over individual variables, and $\mathbf{\Pi}^{\mu \to o}$, for quantification over modal propositional variables that depend on worlds. They become terms of type $(\iota \to (\mu \to o)) \to (\mu \to o)$ and $((\mu \to o) \to (\mu \to o)) \to (\mu \to o)$ respectively.

The $\mathcal{QML}^{STT}$ modal operators $\neg$, $\vee$, $\square$, $\mathbf{\Pi}^{\iota}$, and $\mathbf{\Pi}^{\mu \to o}$ are now simply defined as follows:

$$\neg_{(\mu \to o) \to (\mu \to o)} = \lambda \phi_{\mu \to o \bullet} \lambda W_{\mu \bullet} \neg (\phi\, W)$$

$$\vee_{(\mu \to o) \to (\mu \to o) \to (\mu \to o)} = \lambda \phi_{\mu \to o \bullet} \lambda \psi_{\mu \to o \bullet} \lambda W_{\mu \bullet} \phi\, W \vee \psi\, W$$

$$\square_{(\mu \to \mu \to o) \to (\mu \to o) \to (\mu \to o)} = \lambda R_{\mu \to \mu \to o \bullet} \lambda \phi_{\mu \to o \bullet} \lambda W_{\mu \bullet} \forall V_{\mu \bullet} \neg (R\, W\, V) \vee \phi\, V$$

$$\mathbf{\Pi}^{\iota}_{(\iota \to (\mu \to o)) \to (\mu \to o)} = \lambda \phi_{\iota \to (\mu \to o) \bullet} \lambda W_{\mu \bullet} \forall X_{\iota \bullet} \phi\, X\, W$$

$$\mathbf{\Pi}^{\mu \to o}_{((\mu \to o) \to (\mu \to o)) \to (\mu \to o)} = \lambda \phi_{(\mu \to o) \to (\mu \to o) \bullet} \lambda W_{\mu \bullet} \forall P_{\mu \to o \bullet} \phi\, P\, W$$

Further operators can be introduced as usual, for example, $\top = \lambda W_{\mu \bullet} \top$, $\bot = \neg \top$, $\wedge = \lambda \phi, \psi_{\bullet} \neg (\neg \phi \vee \neg \psi)$, $\supset = \lambda \phi, \psi_{\bullet} \neg \phi \vee \psi$, $\Leftrightarrow = \lambda \phi, \psi_{\bullet} (\phi \supset \psi) \wedge (\psi \supset \phi)$, $\diamond = \lambda R, \phi_{\bullet} \neg (\square R (\neg \phi))$, $\mathbf{\Sigma}^{\iota} = \lambda \phi_{\bullet} \neg \mathbf{\Pi}^{\iota}(\lambda X_{\bullet} \neg \phi\, X)$, $\mathbf{\Sigma}^{\mu \to o} = \lambda \phi_{\bullet} \neg \mathbf{\Pi}^{\mu \to o}(\lambda P_{\bullet} \neg \phi\, P)$.

For defining $\mathcal{QML}^{STT}$-propositions we fix a set $\mathcal{IV}^{STT}$ of individual variables of type $\iota$, a set $\mathcal{PV}^{STT}$ of propositional variables[3] of type $\mu \to o$, and a set $\mathcal{SYM}^{STT}$ of $n$-ary (curried) predicate constants of types $\underbrace{\iota \to \ldots \to \iota}_{n} \to (\mu \to o)$. Moreover, we fix a set $\mathcal{S}^{STT}$ of accessibility relation constants of type $\mu \to \mu \to o$. $\mathcal{QML}^{STT}$-propositions are now defined as the smallest set of $\mathcal{STT}$-terms for which the following hold:

- if $P \in \mathcal{PV}^{STT}$, then $P \in \mathcal{QML}^{STT}$

- if $X^j \in \mathcal{IV}^{STT}$ ($j = 1, \ldots, n$) and $k \in \mathcal{SYM}^{STT}$, then $(k\, X^1 \ldots X^n) \in \mathcal{QML}^{STT}$

- if $\phi, \psi \in \mathcal{QML}^{STT}$, then $\neg \phi \in \mathcal{QML}^{STT}$ and $\phi \vee \psi \in \mathcal{QML}^{STT}$

- if $r \in \mathcal{S}^{STT}$ and $\phi \in \mathcal{QML}^{STT}$, then $\square\, r\, \phi \in \mathcal{QML}^{STT}$.

- if $X \in \mathcal{IV}^{STT}$ and $\phi \in \mathcal{QML}^{STT}$, then $\mathbf{\Pi}^{\iota}(\lambda X_{\bullet} \phi) \in \mathcal{QML}^{STT}$

- if $P \in \mathcal{PV}^{STT}$ and $\phi \in \mathcal{QML}^{STT}$, then $\mathbf{\Pi}^{\mu \to o}(\lambda P_{\bullet} \phi) \in \mathcal{QML}^{STT}$

We write $\square_r \phi$ for $\square\, r\, \phi$, $\forall X_{\iota \bullet} \phi$ for $\mathbf{\Pi}^{\iota}(\lambda X_{\iota \bullet} \phi)$, and $\forall P_{\mu \to o \bullet} \phi$ for $\mathbf{\Pi}^{\mu \to o}(\lambda P_{\mu \to o \bullet} \phi)$.

Note that the defining equations for our $\mathcal{QML}$ modal operators are themselves formulas in simple type theory. Hence, we can express $\mathcal{QML}$ formulas in a HO-ATP elegantly in the usual syntax. For example, $\square_r \exists P_{\mu \to o \bullet} P$ is a $\mathcal{QML}^{STT}$ proposition; it has type $\mu \to o$.

Validity of $\mathcal{QML}^{STT}$ propositions is defined in the obvious way: a $\mathcal{QML}$-proposition $\phi_{\mu \to o}$ is valid if and only if for all possible worlds $w_{\mu}$ we have $w \in \phi_{\mu \to o}$, that is, if and only if $\phi_{\mu \to o}\, w_{\mu}$ holds. Hence, the notion of validity is modeled via the following equation:

$$\text{mvalid} = \lambda \phi_{\mu \to o \bullet} \forall W_{\mu \bullet} \phi\, W$$

---

[3]Note that the denotation of propositional variables depends on worlds.

Now we can formulate proof problems in $\mathcal{QML}^{STT}$, e.g., mvalid $\Box_r \exists P_{\mu \to o}.\, P$. Using rewriting or definition expanding, we can reduce such proof problems to corresponding statements containing only the basic connectives $\neg$, $\vee$, $=$, $\Pi^\iota$, and $\Pi^{\mu \to o}$ of $\mathcal{STT}$. In contrast to the many other approaches no external transformation mechanism is required. For our example we get $\forall W_\mu.\forall Y_\mu.\,\neg(r\,W\,Y) \vee (\neg \forall X_{\mu \to o}.\,\neg(X\,Y))$. It is easy to check that this term is valid in Henkin semantics: put $X = \lambda Y_\mu.\,\top$.

We have proved soundness and completeness for this embedding [10].

**Theorem 2.1 (Soundness)**  *If $\models^{\mathcal{STT}} (\text{valid } s_{\mu \to o})$ then $\models^{\mathbf{QK}\pi} s$.*

**Theorem 2.2 (Completeness)**  *If $\models^{\mathbf{QK}\pi} s$ then $\models^{\mathcal{STT}} (\text{valid } s_{\mu \to o})$.*

These results illustrate the natural correspondence between $\mathbf{QK}\pi$ models and Henkin models. Moreover, we get the following corollaries.

**Corollary 2.3** *The reduction of our embedding to propositional quantified multimodal logics (which only allow quantification over propositional variables) is sound and complete.*

**Corollary 2.4** *The reduction of our embedding to first-order multimodal logics (which only allow quantification over individual variables) is sound and complete.*

**Corollary 2.5** *The reduction of our embedding to propositional multimodal logics (no quantification) is sound and complete.*

In the remainder of the paper we will usually omit type information. It is sufficient to remember that states are of type $\mu$, multimodal propositions of type $\mu \to o$, and accessibility relations of type $\mu \to \mu \to o$.

# 3 A Case Study

In this section we apply off the shelf reasoning systems for simple type theory to simple problems *in* and *about* quantified multimodal logics. Many of these examples have been adapted from Goldblatt's textbook [18]. Our problems are encoded in the new TPTP THF syntax [11] and our experiments exploit the new higher-order TPTP infrastructure [24]. The reasoning systems we apply are LEO-II (version 0.99a) [12], TPS (version 3.0) [1] and IsabelleP (version 2008) and IsabelleM (version 2008) of the Isabelle proof assistant [22].

LEO-II is a resolution based higher-order ATP system. LEO-II is implemented in Objective Caml, and is freely available under a BSD-like license. LEO-II is designed to cooperate with specialist systems for fragments of higher-order logic. Currently, LEO-II is capable of cooperating with the first-order ATP systems E, SPASS, and Vampire. LEO-II directly parses THF0 input and communicates with the cooperating first-order ATP system using TPTP standards.

TPS is a higher-order theorem proving system that has been developed under the supervision of Peter B. Andrews since the 1980s. Theorems can be proved either interactively or automatically. In TPS there are flags that can be set to affect the behavior of automated search. The automated TPS used for solving THF problems uses two different collections of flags, in modes called MS98-FO-MODE and BASIC-MS04-2-MODE. As the two modes have quite different capabilities, they are run in competition parallel as a simple way of obtaining greater coverage.

Isabelle is normally used interactively. A fully automatic version, called IsabelleP, has been implemented using strategy scheduling of the nine automatic tactics simp, blast, auto, metis, fast, fastsimp, best, force, and meson. While it was probably never intended to use Isabelle as a fully automatic system, this simple automation provides useful capability. The ability of Isabelle to find (counter-) models using the refute tactic has also been integrated into an automatic system, called IsabelleM.

The first interesting question clearly is whether our embedding $\mathcal{QML}^{STT}$ of quantified multimodal logic in simple type theory is consistent. The file in Attachment A contains the encoding of our embedding in TPTP THF syntax. The file also contains some further notions and concepts as will be introduced later in this paper. When applying the model generator IsabelleM to this input file, we receive in 0.7 seconds the answer that this set of definitions and axioms is satisfiable. Thus, IsabelleM confirms that our theory is consistent.

All experiments were conducted with the SystemOnTPTP interface [23] which provides online access to the theorem provers running locally at computers at University of Miami. These provers were called remotely with a timeout of 200s.

| Problem | IsabelleP | LEO-II | TPS | IsabelleM |
|---------|-----------|--------|-----|-----------|
| (1) | THM(40.0) | THM(0.0) | THM(0.3) | TMO |
| (2) | THM(105.6) | THM(0.0) | THM(0.4) | TMO |
| (3) | THM(149.6) | THM(0.0) | THM(0.2) | TMO |
| (4) | THM(123.8) | THM(0.0) | THM(0.3) | TMO |
| (5) | TMO | THM(0.1) | THM(0.3) | TMO |
| (6) | TMO | THM(0.1) | THM(0.2) | TMO |

Table 1:

## 3.1 Reasoning within Multimodal Logics

We study whether some simple textbook examples can be automatically proved.

**Example 3.1 (cf. [18], Exercise 1.4)** The following statements are valid (for all accessibility relations $r$).

$$\Box_r \top \tag{1}$$
$$\forall A, B.\ \Box_r (A \supset B) \supset (\Box_r A \supset \Box_r B) \tag{2}$$
$$\forall A, B.\ \Diamond_r (A \supset B) \supset (\Box_r A \supset \Diamond_r B) \tag{3}$$
$$\forall A, B.\ \Box_r (A \supset B) \supset (\Diamond_r A \supset \Diamond_r B) \tag{4}$$
$$\forall A, B.\ \Box_r (A \wedge B) \Leftrightarrow (\Box_r A \wedge \Box_r B) \tag{5}$$
$$\forall A, B.\ \Diamond_r (A \vee B) \Leftrightarrow (\Diamond_r A \vee \Diamond_r B) \tag{6}$$

The THF encoding of Example (3) is given in Appendix B.

The results of our experiment are given in Table 1. The table entries are to be interpreted as follows: THM(40.0) says that a problem was classified as 'theorem' within 40.0 seconds. TMO says that the prover was killed due to reaching the 'timeout' limit of 200 seconds. UKN specifies that the prover stopped proof search but without generating any classification result. CSA(0.8) (cf. Table 2) says that a problem was classified as counter-satisfiable in 0.8 seconds.

The next set of examples is more challenging. These examples are about counter-satisfiable statements (in basic multimodal logic K). Proving them is non-trivial since the synthesis of appropriate accessibility relations $R$ and propositions $A, B$ are required.[4]

---

[4] These examples are actually already belonging to the next subsection, since they rather illustrate the potential of our systems for reasoning about multimodal logics than reasoning within them — the synthesis of concrete accessibility relations, for instance, is to the best of our knowledge not supported in traditional, direct approaches.

**Example 3.2 (cf. [18], Exercise 1.5)** The following propositions are not valid.

$$\forall A. \Box_r A \supset A \tag{7}$$
$$\forall A. \Box_r A \supset \Box_r \Box_r A \tag{8}$$
$$\forall A, B. \Box_r (A \supset B) \supset (\Box_r A \supset \Diamond_r B) \tag{9}$$
$$\Diamond_r \top \tag{10}$$
$$\forall A. \Diamond_r A \supset \Box_r A \tag{11}$$
$$\forall A, B. \Box_r (\Box_r A \supset B) \vee \Box_r (\Box_r B \supset A) \tag{12}$$
$$\forall A, B. \Box_r (A \vee B) \supset (\Box_r A \vee \Box_r B) \tag{13}$$
$$\forall A. \Box_r (\Box_r A \supset A) \supset \Box_r A \tag{14}$$

The encoding of these examples is analogous to the encoding of the previous ones. However, for each problem we also formulate a second question which asks whether there is an accessibility relation $r$ such that the proposition is not valid. For instance, for Example 8 this second question, called problem (8a) below, is encoded as follows:

```
%---- include the definitions for quantified multimodal logic
include('QML.ax').
%---- conjecture statement
thf(conj,conjecture,(
    ? [R: mu > mu > $o] :
      ~ ( mvalid
          @ ( mall_prop
              @ ^ [A: mu > $o] :
                ( mimpl @ ( mbox @ R @ A )
                        @ ( mbox @ R @ ( mbox @ R @ A ) ) ) ) )).
```

The results of this experiment are given in Table 2. Note that our provers generate the expected results for examples (7),(7a),(9),(9a),(10),(10a),(14), and (14a), that is, the model finder IsabelleM finds a countermodel for Examples (7),(9),(10), and (14) and even the provers can show that there exists countermodels by proving the corresponding statements (7a),(9a),(10a), and (14a).

For the remaining problems the task is more challenging though and we will illustrate this with the help of Example (8), the transitivity axiom. This axiom is not valid and correctly none of the theorem provers signals that a proof can be found. Unfortunately, the model generator IsabelleM does not generate a countermodel either. We thus try to prove (8a) which states that there is an accessibility relation $R$ such that the transitivity axiom is invalidated.

$$\exists R. (\neg\text{valid } \forall A. (\Box_R A \supset \Box_R \Box_R A))$$

We would expect that this statement can be quickly proved when instantiating variable $R$ with a non-transitive relation. For good reasons our provers fail to do so. In fact, our provers also fail to prove the related statement

$$\exists R. \neg(\text{transitive } R)$$

| Problem | IsabelleP | LEO-II | TPS | IsabelleM |
|---------|-----------|----------|----------|-----------|
| (7)     | TMO       | UKN      | TMO      | CSA(1.6)  |
| (7a)    | THM(79.6) | THM(62.4)| THM(0.2) | TMO       |
| (8)     | UKN       | UKN      | TMO      | TMO       |
| (8a)    | UKN       | TMO      | TMO      | CSA(1.6)  |
| (9)     | UKN       | TMO      | TMO      | CSA(1.7)  |
| (9a)    | UKN       | THM(64.9)| THM(0.2) | TMO       |
| (10)    | UKN       | UKN      | TMO      | CSA(1.5)  |
| (10a)   | THM(78.9) | THM(63.4)| THM(0.2) | TMO       |
| (11)    | UKN       | UKN      | TMO      | TMO       |
| (11a)   | TMO       | TMO      | UKN      | CSA(1.6)  |
| (12)    | TMO       | UKN      | TMO      | TMO       |
| (12a)   | UKN       | TMO      | TMO      | CSA(1.8)  |
| (13)    | UKN       | UKN      | TMO      | TMO       |
| (13a)   | UKN       | TMO      | TMO      | CSA(1.7)  |
| (14)    | UKN       | TMO      | TMO      | CSA(1.6)  |
| (14a)   | UKN       | THM(72.5)| THM(8.6) | TMO       |

Table 2:

The reason is that without further assumptions this statement is not a theorem. We have neither assumed the axiom of infinity nor that there exist at least two different possible worlds. Hence, our domain of possible worlds may well just consist of a single world $w$ in which case a non-transitive accessibility relation cannot be provided. It is thus not surprising that IsabelleM finds a countermodel to statement (8a). Unfortunately, it does not succeed though to find a countermodel to the original problem.

## 3.2 Reasoning about Multimodal Logics

The literature on modal logics is full of theorems and student exercises illustrating the relationships between properties of accessibility relations and corresponding axioms (respectively axiom schemata). Therefore, the question that interests us next is whether such meta-theoretic results can be automatically proved by our higher-order reasoners. We again study the examples given in Goldblatt's textbook [18]. First we encode various accessibility relation properties in simple type theory.

$$\text{reflexive} = \lambda R.\forall S.\, R\, S\, S \tag{15}$$

$$\text{symmetric} = \lambda R.\forall S, T.\, (R\, S\, T) \Rightarrow (R\, T\, S) \tag{16}$$

$$\text{serial} = \lambda R.\forall S.\exists T.\, (R\, S\, T) \tag{17}$$

$$\text{transitive} = \lambda R.\forall S, T, U.\, ((R\, S\, T) \wedge (R\, T\, U)) \Rightarrow (R\, S\, U) \tag{18}$$

$$\text{Euclidean} = \lambda R.\forall S, T, U.\, ((R\, S\, T) \wedge (R\, S\, U)) \Rightarrow (R\, T\, U) \tag{19}$$

$$\text{partially\_functional} = \lambda R.\forall S, T, U.\, ((R\, S\, T) \wedge (R\, S\, U)) \Rightarrow (T = U) \tag{20}$$

$$\text{functional} = \lambda R.\forall S.\exists T.\, (R\, S\, T) \wedge \forall U.\, (R\, S\, U) \Rightarrow (T = U) \tag{21}$$

$$\text{weakly\_dense} = \lambda R.\forall S, T.\, (R\, S\, T) \Rightarrow \exists U.\, (R\, S\, U) \wedge (R\, U\, T) \tag{22}$$

$$\text{weakly\_connected} = \lambda R.\forall S, T, U.\, ((R\, S\, T) \wedge (R\, S\, U)) \Rightarrow\\ ((R\, T\, U) \vee (T = U) \vee (R\, U\, T)) \tag{23}$$

$$\text{weakly\_directed} = \lambda R.\forall S, T, U.\, ((R\, S\, T) \wedge (R\, S\, U)) \Rightarrow\\ \exists V.\, (R\, T\, V) \wedge (R\, U\, V) \tag{24}$$

The formulation of these accessibility relation properties in THF syntax is straightforward, for example,[5]

```
thf(mtransitive,definition,
  ( mtransitive
  = ( ^ [R: mu > mu > $o]: ! [S: mu,T: mu,U: mu]:
       (((R @ S @ T) & (R @ T @ U)) => (R @ S @ U))))).
```

The corresponding axioms are given next.

$$\forall A.\, \Box_r A \supset A \tag{25}$$

$$\forall A.\, A \supset \Box_r \Diamond_r A \tag{26}$$

$$\forall A.\, \Box_r A \supset \Diamond_r A \tag{27}$$

$$\forall A.\, \Box_r A \supset \Box_r \Box_r A \tag{28}$$

$$\forall A.\, \Diamond_r A \supset \Box_r \Diamond_r A \tag{29}$$

$$\forall A.\, \Diamond_r A \supset \Box_r A \tag{30}$$

$$\forall A.\, \Diamond_r A \Leftrightarrow \Box_r A \tag{31}$$

$$\forall A.\, \Box_r \Box_r A \supset \Box_r A \tag{32}$$

$$\forall A, B.\, \Box_r (A \wedge (\Box_r A \supset B)) \vee \Box_r (B \wedge (\Box_r B \supset A)) \tag{33}$$

$$\forall A.\, \Diamond_r \Box_r A \supset \Box_r \Diamond_r A \tag{34}$$

---

[5]We have decided to use a prefix 'm' for all these example to avoid clashes of these definitions with inbuilt concepts in our reasoners using the same name.

| Problem | IsabelleP | LEO-II | TPS | IsabelleM |
|---|---|---|---|---|
| (15) $\Rightarrow$ (25) | THM(90.0) | THM(0.0) | THM(0.3) | TMO |
| (16) $\Rightarrow$ (26) | THM(113.7) | THM(0.0) | THM(0.2) | TMO |
| (17) $\Rightarrow$ (27) | THM(102.1) | THM(0.0) | THM(0.3) | TMO |
| (18) $\Rightarrow$ (28) | THM(123.2) | THM(0.0) | THM(8.1) | TMO |
| (19) $\Rightarrow$ (29) | THM(131.7) | THM(0.0) | THM(0.6) | TMO |
| (20) $\Rightarrow$ (30) | THM(125.8) | THM(0.0) | THM(1.1) | TMO |
| (21) $\Rightarrow$ (31) | THM(156.3) | THM(0.0) | TMO | TMO |
| (22) $\Rightarrow$ (32) | THM(105.1) | THM(0.0) | THM(1.1) | TMO |
| (23) $\Rightarrow$ (33) | THM(42.6) | THM(0.0) | THM(0.2) | TMO |
| (24) $\Rightarrow$ (34) | THM(178.1) | THM(0.0) | THM(0.2) | TMO |
| (15) $\Leftarrow$ (25) | THM(75.3) | THM(0.0) | THM(0.2) | TMO |
| (16) $\Leftarrow$ (26) | UKN | TMO | THM(0.5) | TMO |
| (17) $\Leftarrow$ (27) | THM(89.6) | THM(0.0) | THM(0.2) | TMO |
| (18) $\Leftarrow$ (28) | TMO | TMO | TMO | TMO |
| (19) $\Leftarrow$ (29) | UKN | TMO | TMO | TMO |
| (20) $\Leftarrow$ (30) | TMO | THM(6.2) | THM(47.9) | TMO |
| (21) $\Leftarrow$ (31) | TMO | TMO | THM(151.3) | TMO |
| (22) $\Leftarrow$ (32) | TMO | TMO | TMO | TMO |
| (23) $\Leftarrow$ (33) | UKN | TMO | TMO | TMO |
| (24) $\Leftarrow$ (34) | THM(85.1) | THM(0.1) | THM(148.5) | TMO |

Table 3:

**Example 3.3 (Correspondence of accessibility relation properties and axioms)**
Property $k$ ($k = (15), \ldots, (24)$) holds for accessibility relation $r$ if and only if the corresponding axiom (k+10) is valid.

As an example we present the encoding of problem (18) $\Leftarrow$ (28) in Appendix C. The outcome of this experiment is summarized in Table 3. Except for 4 out of the 20 problems our provers collectively are able to come up with the expected results.

## 3.3 Epistemic Reasoning

In this section we encode epistemic reasoning problems in our framework. The examples are taken from Baldoni [5].

**Example 3.4 (Epistemic reasoning: The friends puzzle)**   (i) Peter is a friend of John, so if Peter knows that John knows something then John knows that Peter knows the same thing. (ii) Peter is married, so if Peter's wife knows something, then Peter knows the same thing. John and Peter have an appointment, let us consider the following situation: (a) Peter knows the time of their appointment. (b) Peter also knows that John knows the place of their appointment. Moreover, (c) Peter's wife knows that if Peter knows the time of their appointment, then John knows that too (since John and Peter are friends). Finally, (d) Peter knows that if John knows the place and the time of their appointment, then John knows that he has an appointment.

From this situation we want to prove (e) that each of the two friends knows that the other one knows that he has an appointment.

For the three persons in this puzzle, John, Peter and Peter's wife, we introduce corresponding accessibility relations *john*, *peter*, and *wife(peter)*. For this example it is sufficient to require the S4 axioms, that is, a knowledge or truth axioms (cf. 25) and positive introspection axioms (cf. 28) for each accessibility relation. Alternatively, we here require that the accessibility relations *john*, *peter*, and *wife(peter)* are reflexive and transitive.

```
%----
thf(peter,type,( peter: mu > mu > $o )).
thf(john,type,( john: mu > mu > $o )).
thf(wife,type,(wife: ( mu > mu > $o ) > mu > mu > $o )).
%----
thf(refl_peter,axiom,( mreflexive @ peter )).
thf(refl_john,axiom,( mreflexive @ john )).
thf(refl_wife_peter,axiom,( mreflexive @ ( wife @ peter ))).
thf(trans_peter,axiom,( mtransitive @ peter )).
thf(trans_john,axiom,( mtransitive @ john)).
thf(trans_wife_peter,axiom,( mtransitive @ ( wife @ peter ))).
```

It is clear that the following S4 axioms are implied by these conditions on the accessibility relations ($i$ ranges over *wife(peter)*, *peter*, and *john*).

$$K(i) = \forall A.\, \Box_i A \supset A \qquad T(i) = \forall A.\, \Box_i A \supset \Box_i \Box_i A$$

We again use our reasoners to verify this. We only check this for K(wife(peter)) and T(wife(peter)); the results are given in Table 4.

Next, we encode the facts from the puzzle. For (i) we provide a persistence axiom and for (ii) an inclusion axiom:

$$\forall A.\, \Box_{\text{peter}} \Box_{\text{john}} A \supset \Box_{\text{john}} \Box_{\text{peter}} A \tag{35}$$

$$\forall A.\, \Box_{\text{wife(peter)}} A \supset \Box_{\text{peter}} A \tag{36}$$

| Problem | IsabelleP | LEO-II | TPS | IsabelleM |
|---|---|---|---|---|
| K(wife(peter)) | THM(89.4) | THM(0.0) | THM(0.3) | TMO |
| T(wife(peter)) | THM(178.3) | THM(0.0) | TMO | TMO |
| Example 3.4 | TMO | THM(0.1) | TMO | TMO |
| Example 3.5 | UKN | THM(0.3) | TMO | TMO |

Table 4:

Finally, the facts (a)-(d) and the conclusion (e) are encoded as follows (time, place, and appointment are propositional constants, that is, constants of type $\iota \to o$ in our framework):

$$\Box_{peter} \text{ time} \tag{37}$$

$$\Box_{peter} \Box_{john} \text{ place} \tag{38}$$

$$\Box_{wife(peter)} (\Box_{peter} \text{ time} \supset \Box_{john} \text{ time}) \tag{39}$$

$$\Box_{peter} \Box_{john} (\text{place} \wedge \text{time} \supset \text{appointment}) \tag{40}$$

$$\Box_{john} \Box_{peter} \text{ appointment} \wedge \Box_{peter} \Box_{john} \text{ appointment} \tag{41}$$

The THF encoding of the entire example is presented in Appendix D. Table 4 shows the results of our experiment.

In the modeling of our next example we also follow Baldoni.

**Example 3.5 (Wise men puzzle)** Once upon a time, a king wanted to find the wisest out of his three wisest men. He arranged them in a circle and told them that he would put a white or a black spot on their foreheads and that one of the three spots would certainly be white. The three wise men could see and hear each other but, of course, they could not see their faces reflected anywhere. The king, then, asked to each of them to find out the color of his own spot. After a while, the wisest correctly answered that his spot was white.

The encoding of the example is presented in Appendix E and the performance of our provers shown in Table 4. For explanations on the modeling of this example we refer to [5] pp. 55-57.

# 4  Outlook

This work presented in this paper has its roots in the LEO-II project (in 2006/2007 at Cambridge University) in which we first studied and employed the presented embedding of quantified multimodal logics in simple type theory [8][6]. Subsequently, we have applied the idea also to propositional intuitionistic logics and to access control logics [6, 9]. And similar results are possible for various other non-classical logics that have been discussed in the literature.

For example, to model temporal logic in our framework, we may fix particular accessibility relations relations *past* and *future*. Moreover, we require that both relations are transitive and mutually inverse. This can be easily done by stating:

$$(\text{transitive past}) \wedge (\text{transitive future}) \tag{42}$$
$$\forall S, T_{\bullet} (\text{past } S\ T) \Leftrightarrow (\text{future } T\ S) \tag{43}$$
$$\tag{44}$$

We may then introduce further temporal operators, for example, *always* and *sometime* (both in the past, at present, and in the future):

$$\text{always} = \lambda A_{\bullet} \Box_{\text{past}} A \ \wedge \ A \ \wedge \ \Box_{\text{future}} A \tag{45}$$
$$\text{sometime} = \lambda A_{\bullet} \Diamond_{\text{past}} A \ \vee \ A \ \vee \ \Diamond_{\text{future}} A \tag{46}$$

The exploration of (quantified) temporal logic is future work. Future work also includes further extensions of our embedding to also cover quantified hybrid logics [13, 14] and full higher-order modal logics [17, 20]. A first suggestion in direction of higher-order modal logics has already been made [8]. This proposal does however not yet address intensionality aspects. However, combining this proposal with non-extensional notions of models for simple type theory [7, 21] appears a promising direction.

Our overall goal is to show that various interesting non-classical logics can be fruitfully mechanized and partly automated in modern proof assistants with the help of our embedding. We also want to motivate the integration of our automatic higher-order reasoners in modern proof assistants as well as the development of further automated reasoning tools for classical higher-order logic. It is obvious that the existing reasoners should be significantly improved for fruitful application to challenge problems in practice. This seems not unreasonable though. Moreover, when working with our reasoners from within a proof assistant then the user may provide interactive help, for example, by formulating some lemmas or by splitting proof tasks in simpler subtasks. In this context it seems also useful that our reasoners produce proof objects in a standard proof representation format, e.g., in the TSTP format, and that a translation from this format into the proof representations used in the proof assistants is provided.

Finally, it may be possible to formally verify the entire theory of our embedding within a proof assistant.

---

[6]This paper was written in 2007.

# A  Modeling quantified multimodal logic in simple type theory

```
%------------------------------------------------------------------
% File     : QML.ax
% Domain   : Quantified multimodal logic
% Problems :
% Version  :
% English  : Embedding of quantified multimodal logic in
%            simple type theory
% Refs     :
% Source   : Formalization in THF by C. Benzmueller
% Names    :
% Status   :
% Rating   :
% Syntax   :
% Comments :
%------------------------------------------------------------------
%---- declaration of additional base type mu
thf(mu,type,(
    mu: $tType )).

%---- modal operators not, or, box, Pi
thf(mnot,definition,
    ( mnot
    = ( ^ [Phi: mu > $o,W: mu] :
          ~ ( Phi @ W ) ) )).

thf(mor,definition,
    ( mor
    = ( ^ [Phi: mu > $o,Psi: mu > $o,W: mu] :
          ( ( Phi @ W )
          | ( Psi @ W ) ) ) )).

thf(mbox,definition,
    ( mbox
    = ( ^ [R: mu > mu > $o,Phi: mu > $o,W: mu] :
        ! [V: mu] :
          ( ~ ( R @ W @ V )
          | ( Phi @ V ) ) ) )).

thf(mall_ind,definition,
    ( mall_ind
    = ( ^ [Phi: $i > mu > $o,W: mu] :
        ! [X: $i] :
          ( Phi @ X @ W ) ) )).

thf(mall_prop,definition,
    ( mall_prop
    = ( ^ [Phi: ( mu > $o ) > mu > $o,W: mu] :
        ! [P: mu > $o] :
          ( Phi @ P @ W ) ) )).

%---- further modal operators
thf(mtrue,definition,
    ( mtrue
    = ( ^ [W: mu] : $true ) )).
```

```
thf(mtrue,definition,
    ( mfalse
    = ( mall_prop
      @ ^ [P: mu > $o] :
          ( mnot @ mtrue ) ) )).

thf(mand,definition,
    ( mand
    = ( ^ [Phi: mu > $o,Psi: mu > $o] :
          ( mnot @ ( mor @ ( mnot @ Phi ) @ ( mnot @ Psi ) ) ) ) )).

thf(mimpl,definition,
    ( mimpl
    = ( ^ [Phi: mu > $o,Psi: mu > $o] :
          ( mor @ ( mnot @ Phi ) @ Psi ) ) )).

thf(mequiv,definition,
    ( mequiv
    = ( ^ [Phi: mu > $o,Psi: mu > $o] :
          ( mand @ ( mimpl @ Phi @ Psi )
               @ ( mimpl @ Psi @ Phi ) ) ) )).

thf(mdia,definition,
    ( mdia
    = ( ^ [R: mu > mu > $o,Phi: mu > $o] :
          ( mnot @ ( mbox @ R @ ( mnot @ Phi ) ) ) ) )).

thf(mexi_ind,definition,
    ( mexi_ind
    = ( ^ [Phi: $i > mu > $o] :
          ( mnot
          @ ( mall_ind
            @ ^ [X: $i] :
                ( mnot @ ( Phi @ X ) ) ) ) ) )).

thf(mexi_prop,definition,
    ( mexi_prop
    = ( ^ [Phi: ( mu > $o ) > mu > $o] :
          ( mnot
          @ ( mall_prop
            @ ^ [P: mu > $o] :
                ( mnot @ ( Phi @ P ) ) ) ) ) )).

%----------------------------------------------------------------
%---- temporal logic

thf(past,type,(
    past: mu > mu > $o )).

thf(future,type,(
    future: mu > mu > $o )).

thf(past_transitive,axiom,
    ( mtransitive @ past )).
```

```
thf(future_transitive,axiom,
    ( mtransitive @ future )).

thf(past_future,axiom,(
    ! [S: mu,T: mu] :
      ( ( future @ S @ T )
    <=> ( past @ T @ S ) ) )).

%---- definition of always
thf(malways,definition,
    ( malways
    = ( ^ [A: mu > $o] :
          ( mand @ ( mand @ ( mbox @ past @ A ) @ A )
              @ ( mbox @ future @ A ) ) ) )).

%---- definition of sometime
thf(msometime,definition,
    ( msometime
    = ( ^ [A: mu > $o] :
          ( mor @ ( mor @ ( mdia @ past @ A ) @ A )
              @ ( mdia @ future @ A ) ) ) )).

%------------------------------------------------------------------
%---- definition of validity
thf(mvalid,definition,
    ( mvalid
    = ( ^ [Phi: mu > $o] :
        ! [W: mu] :
          ( Phi @ W ) ) )).

%---- definition of satisfiability
thf(msatisfiable,definition,
    ( msatisfiable
    = ( ^ [Phi: mu > $o] :
        ? [W: mu] :
          ( Phi @ W ) ) )).

%---- definition of countersatisfiability
thf(mcountersatisfiable,definition,
    ( mcountersatisfiable
    = ( ^ [Phi: mu > $o] :
        ? [W: mu] :
          ~ ( Phi @ W ) ) )).

%---- definition of invalidity
thf(minvalid,definition,
    ( minvalid
    = ( ^ [Phi: mu > $o] :
        ! [W: mu] :
          ~ ( Phi @ W ) ) )).

%------------------------------------------------------------------
%---- definition of properties of accessibility relations
thf(mreflexive,definition,
    ( mreflexive
```

```
        = ( ^ [R: mu > mu > $o] :
            ! [S: mu] :
              ( R @ S @ S ) ) )).

thf(msymmetric,definition,
    ( msymmetric
    = ( ^ [R: mu > mu > $o] :
        ! [S: mu,T: mu] :
          ( ( R @ S @ T )
         => ( R @ T @ S ) ) ) )).

thf(mserial,definition,
    ( mserial
    = ( ^ [R: mu > mu > $o] :
        ! [S: mu] :
        ? [T: mu] :
          ( R @ S @ T ) ) )).

thf(mtransitive,definition,
    ( mtransitive
    = ( ^ [R: mu > mu > $o] :
        ! [S: mu,T: mu,U: mu] :
          ( ( ( R @ S @ T )
            & ( R @ T @ U ) )
         => ( R @ S @ U ) ) ) )).

thf(meuclidean,definition,
    ( meuclidean
    = ( ^ [R: mu > mu > $o] :
        ! [S: mu,T: mu,U: mu] :
          ( ( ( R @ S @ T )
            & ( R @ S @ U ) )
         => ( R @ T @ U ) ) ) )).

thf(mpartially_functional,definition,
    ( mpartially_functional
    = ( ^ [R: mu > mu > $o] :
        ! [S: mu,T: mu,U: mu] :
          ( ( ( R @ S @ T )
            & ( R @ S @ U ) )
         => ( T = U ) ) ) )).

thf(mfunctional,definition,
    ( mfunctional
    = ( ^ [R: mu > mu > $o] :
        ! [S: mu] :
        ? [T: mu] :
          ( ( R @ S @ T )
          & ! [U: mu] :
              ( ( R @ S @ U )
             => ( T = U ) ) ) ) )).

thf(mweakly_dense,definition,
    ( mweakly_dense
    = ( ^ [R: mu > mu > $o] :
        ! [S: mu,T: mu,U: mu] :
```

```
            ( ( R @ S @ T )
         => ? [U: mu] :
              ( ( R @ S @ U )
              & ( R @ U @ T ) ) ) ) )).

thf(mweakly_connected,definition,
    ( mweakly_connected
    = ( ^ [R: mu > mu > $o] :
        ! [S: mu,T: mu,U: mu] :
          ( ( ( R @ S @ T )
            & ( R @ S @ U ) )
         => ( ( R @ T @ U )
            | ( T = U )
            | ( R @ U @ T ) ) ) )).

thf(mweakly_directed,definition,
    ( mweakly_directed
    = ( ^ [R: mu > mu > $o] :
        ! [S: mu,T: mu,U: mu] :
          ( ( ( R @ S @ T )
            & ( R @ S @ U ) )
         => ? [V: mu] :
              ( ( R @ T @ V )
              & ( R @ U @ V ) ) ) )).
```

# B  A simple problem encoding

```
%----------------------------------------------------------------
% File     :
% Domain   :
% Problems :
% Version  :
% English  :
% Refs     :
% Source   : Formalization in THF by C. Benzmueller
% Names    :
% Status   :
% Rating   :
% Syntax   :
% Comments :
%----------------------------------------------------------------
%---- include the definitions for quantified multimodal logic
include('QML.ax').

%---- conjecture statement
thf(conj,conjecture,(
    ! [R: mu > mu > $o] :
      ( mvalid
      @ ( mall_prop
        @ ^ [A: mu > $o] :
            ( mall_prop
            @ ^ [B: mu > $o] :
                ( mimpl @ ( mdia @ R @ ( mimpl @ A @ B ) )
                        @ ( mimpl @ ( mbox @ R @ A )
                                  @ ( mdia @ R @ B ) ) ) ) ) )).
```

# C Correspondence between accessibility relation properties and modal axioms

```
%--------------------------------------------------------------------
% File     :
% Domain   :
% Problems :
% Version  :
% English  :
% Refs     :
% Source   : Formalization in THF by C. Benzmueller
% Names    :
% Status   :
% Rating   :
% Syntax   :
% Comments :
%--------------------------------------------------------------------
%---- include the definitions for quantified multimodal logic
include('QML.ax').

---- conjecture statement
thf(conj,conjecture,(
    ! [R: mu > mu > $o] :
      ( ( mvalid
        @ ( mall_prop
          @ ^ [A: mu > $o] :
              ( mimpl @ ( mbox @ R @ A )
                     @ ( mbox @ R @ ( mbox @ R @ A ) ) ) ) )
     => ( mtransitive @ R ) ) )).
```

# D   A simple example in epistemic logic

```
%--------------------------------------------------------------------
% File     :
% Domain   :
% Problems :
% Version  :
% English  :
% Refs     :
% Source   : Formalization in THF by C. Benzmueller
% Names    :
% Status   :
% Rating   :
% Syntax   :
% Comments :
%--------------------------------------------------------------------
%---- include the definitions for quantified multimodal logic
include('QML.ax').

%----
thf(peter,type,(
    peter: mu > mu > $o )).

thf(john,type,(
    john: mu > mu > $o )).

thf(wife,type,(
    wife: ( mu > mu > $o ) > mu > mu > $o )).

%----
thf(refl_peter,axiom,
    ( mreflexive @ peter )).

thf(refl_john,axiom,
    ( mreflexive @ john )).

thf(refl_wife_peter,axiom,
    ( mreflexive @ ( wife @ peter ) )).

thf(trans_peter,axiom,
    ( mtransitive @ peter )).

thf(trans_john,axiom,
    ( mtransitive @ john )).

thf(trans_wife_peter,axiom,
    ( mtransitive @ ( wife @ peter ) )).

%----
thf(ax_i,axiom,
    ( mvalid
    @ ( mall_prop
      @ ^ [A: mu > $o] :
          ( mimpl @ ( mbox @ peter @ ( mbox @ john @ A ) )
                  @ ( mbox @ john @ ( mbox @ peter @ A ) ) ) ) )).
```

```
thf(ax_ii,axiom,
    ( mvalid
    @ ( mall_prop
      @ ^ [A: mu > $o] :
          ( mimpl @ ( mbox @ ( wife @ peter ) @ A )
                 @ ( mbox @ peter @ A ) ) ) )).

thf(time,type,(
    time: mu > $o )).

thf(place,type,(
    place: mu > $o )).

thf(appointment,type,(
    appointment: mu > $o )).

thf(ax_a,axiom,
    ( mvalid @ ( mbox @ peter @ time ) )).

thf(ax_b,axiom,
    ( mvalid @ ( mbox @ peter @ ( mbox @ john @ place ) ) )).

thf(ax_c,axiom,
    ( mvalid @ ( mbox @ ( wife @ peter )
                      @ ( mimpl @ ( mbox @ peter @ time )
                               @ ( mbox @ john @ time ) ) ) )).

thf(ax_d,axiom,
    ( mvalid @
      ( mbox @ peter
            @ ( mbox @ john
                    @ ( mimpl @ ( mand @ place @ time )
                             @ appointment ) ) ) )).

thf(conj,conjecture,
    ( mvalid @
      ( mand @ ( mbox @ peter
                      @ ( mbox @ john @ appointment ) )
            @ ( mbox @ john
                    @ ( mbox @ peter @ appointment ) ) ) )).
```

# E  Wise men puzzle

```
%----------------------------------------------------------------
% File     :
% Domain   :
% Problems : Wise men puzzle
% Version  :
% English  :
% Refs     : Matteo Baldoni, Normal Multimodal Logics: Automatic
%            Deduction and Logic Programming Extension, Phd thesis,
%            Universita degli studi di Torino, 2003, pp.55-57
% Source   : Formalization in THF by C. Benzmueller
% Names    :
% Status   :
% Rating   :
% Syntax   :
% Comments :
%----------------------------------------------------------------
%---- include the definitions for quantified multimodal logic
include('QML.ax').

%----
thf(a,type,(
    a: mu > mu > $o )).

thf(b,type,(
    b: mu > mu > $o )).

thf(c,type,(
    c: mu > mu > $o )).

thf(fool,type,(
    fool: mu > mu > $o )).

thf(ws,type,(
    ws: ( mu > mu > $o ) > mu > $o )).

%----
thf(axiom_1,axiom,
( mvalid @ ( mbox @ fool @
  ( mor @ ( ws @ a ) @ ( mor @ ( ws @ b ) @ ( ws @ c ) ) ) ) )).

%----
thf(axiom_2_a_b,axiom,
( mvalid @ ( mbox @ fool @
  ( mimpl @ ( ws @ a ) @ ( mbox @ b @ ( ws @ a ) ) ) ) )).

thf(axiom_2_a_c,axiom,
( mvalid @ ( mbox @ fool @
  ( mimpl @ ( ws @ a ) @ ( mbox @ c @ ( ws @ a ) ) ) ) )).

thf(axiom_2_b_a,axiom,
( mvalid @ ( mbox @ fool @
  ( mimpl @ ( ws @ b ) @ ( mbox @ a @ ( ws @ b ) ) ) ) )).

thf(axiom_2_b_c,
```

```
( mvalid @ ( mbox @ fool @
  ( mimpl @ ( ws @ b ) @ ( mbox @ c @ ( ws @ b ) ) ) ) )).

thf(axiom_2_c_a,axiom,
( mvalid @ ( mbox @ fool @
  ( mimpl @ ( ws @ c ) @ ( mbox @ a @ ( ws @ c ) ) ) ) )).

thf(axiom_2_b_a,axiom,
( mvalid @ ( mbox @ fool @
  ( mimpl @ ( ws @ c ) @ ( mbox @ b @ ( ws @ c ) ) ) ) )).

%----
thf(axiom_3_a_b,axiom,
( mvalid @ ( mbox @ fool @
  ( mimpl @ ( mnot @ ( ws @ a ) )
        @ ( mbox @ b @ ( mnot @ ( ws @ a ) ) ) ) ) )).

thf(axiom_3_a_c,axiom,
( mvalid @ ( mbox @ fool @
  ( mimpl @ ( mnot @ ( ws @ a ) )
        @ ( mbox @ c @ ( mnot @ ( ws @ a ) ) ) ) ) )).

thf(axiom_3_b_a,axiom,
( mvalid @ ( mbox @ fool @
  ( mimpl @ ( mnot @ ( ws @ b ) )
        @ ( mbox @ a @ ( mnot @ ( ws @ b ) ) ) ) ) )).

thf(axiom_3_b_c,axiom,
( mvalid @ ( mbox @ fool @
  ( mimpl @ ( mnot @ ( ws @ b ) )
        @ ( mbox @ c @ ( mnot @ ( ws @ b ) ) ) ) ) )).

thf(axiom_3_c_a,axiom,
( mvalid @ ( mbox @ fool @
  ( mimpl @ ( mnot @ ( ws @ c ) )
        @ ( mbox @ a @ ( mnot @ ( ws @ c ) ) ) ) ) )).

thf(axiom_3_b_a,axiom,
( mvalid @ ( mbox @ fool @
  ( mimpl @ ( mnot @ ( ws @ c ) )
        @ ( mbox @ b @ ( mnot @ ( ws @ c ) ) ) ) ) )).

%----
thf(t_axiom_for_fool,axiom,
    ( mvalid
    @ ( mall_prop
      @ ^ [A: mu > $o] :
          ( mimpl @ ( mbox @ fool @ A ) @ A ) ) )).

thf(k_axiom_for_fool,axiom,
    ( mvalid
    @ ( mall_prop
      @ ^ [A: mu > $o] :
          ( mimpl @ ( mbox @ fool @ A )
                @ ( mbox @ fool @ ( mbox @ fool @ A ) ) ) ) )).
```

```
thf(i_axiom_for_fool_a,axiom,
    ( mvalid
    @ ( mall_prop
      @ ^ [Phi: mu > $o] :
          ( mimpl @ ( mbox @ fool @ Phi )
                 @ ( mbox @ a @ Phi ) ) ) )).

thf(i_axiom_for_fool_b,axiom,
    ( mvalid
    @ ( mall_prop
      @ ^ [Phi: mu > $o] :
          ( mimpl @ ( mbox @ fool @ Phi )
                 @ ( mbox @ b @ Phi ) ) ) )).

thf(i_axiom_for_fool_c,axiom,
    ( mvalid
    @ ( mall_prop
      @ ^ [Phi: mu > $o] :
          ( mimpl @ ( mbox @ fool @ Phi )
                 @ ( mbox @ c @ Phi ) ) ) )).

%----
thf(a7_axiom_for_fool_a_b,axiom,
    ( mvalid
    @ ( mall_prop
      @ ^ [Phi: mu > $o] :
          ( mimpl @ ( mbox @ a @ Phi )
                 @ ( mbox @ b @ ( mbox @ a @ Phi ) ) ) ) )).

thf(a7_axiom_for_fool_a_c,axiom,
    ( mvalid
    @ ( mall_prop
      @ ^ [Phi: mu > $o] :
          ( mimpl @ ( mbox @ a @ Phi )
                 @ ( mbox @ c @ ( mbox @ a @ Phi ) ) ) ) )).

thf(a7_axiom_for_fool_b_a,axiom,
    ( mvalid
    @ ( mall_prop
      @ ^ [Phi: mu > $o] :
          ( mimpl @ ( mbox @ b @ Phi )
                 @ ( mbox @ a @ ( mbox @ b @ Phi ) ) ) ) )).

thf(a7_axiom_for_fool_b_c,axiom,
    ( mvalid
    @ ( mall_prop
      @ ^ [Phi: mu > $o] :
          ( mimpl @ ( mbox @ b @ Phi )
                 @ ( mbox @ c @ ( mbox @ b @ Phi ) ) ) ) )).

thf(a7_axiom_for_fool_c_a,axiom,
    ( mvalid
    @ ( mall_prop
      @ ^ [Phi: mu > $o] :
          ( mimpl @ ( mbox @ c @ Phi )
                 @ ( mbox @ a @ ( mbox @ c @ Phi ) ) ) ) )).
```

```
thf(a7_axiom_for_fool_c_b,axiom,
    ( mvalid
    @ ( mall_prop
      @ ^ [Phi: mu > $o] :
          ( mimpl @ ( mbox @ c @ Phi )
                 @ ( mbox @ b @ ( mbox @ c @ Phi ) ) ) ) )).

%----
thf(a6_axiom_for_fool_a_b,axiom,
    ( mvalid
    @ ( mall_prop
      @ ^ [Phi: mu > $o] :
          ( mimpl @ ( mnot @ ( mbox @ a @ Phi ) )
                 @ ( mbox @ b @ ( mnot @ ( mbox @ a @ Phi ) ) ) ) ) )).

thf(a6_axiom_for_fool_a_c,axiom,
    ( mvalid
    @ ( mall_prop
      @ ^ [Phi: mu > $o] :
          ( mimpl @ ( mnot @ ( mbox @ a @ Phi ) )
                 @ ( mbox @ c @ ( mnot @ ( mbox @ a @ Phi ) ) ) ) ) )).

thf(a6_axiom_for_fool_b_a,axiom,
    ( mvalid
    @ ( mall_prop
      @ ^ [Phi: mu > $o] :
          ( mimpl @ ( mnot @ ( mbox @ b @ Phi ) )
                 @ ( mbox @ a @ ( mnot @ ( mbox @ b @ Phi ) ) ) ) ) )).

thf(a6_axiom_for_fool_b_c,axiom,
    ( mvalid
    @ ( mall_prop
      @ ^ [Phi: mu > $o] :
          ( mimpl @ ( mnot @ ( mbox @ b @ Phi ) )
                 @ ( mbox @ c @ ( mnot @ ( mbox @ b @ Phi ) ) ) ) ) )).

thf(a6_axiom_for_fool_c_a,axiom,
    ( mvalid
    @ ( mall_prop
      @ ^ [Phi: mu > $o] :
          ( mimpl @ ( mnot @ ( mbox @ c @ Phi ) )
                 @ ( mbox @ a @ ( mnot @ ( mbox @ c @ Phi ) ) ) ) ) )).

thf(a6_axiom_for_fool_c_b,axiom,
    ( mvalid
    @ ( mall_prop
      @ ^ [Phi: mu > $o] :
          ( mimpl @ ( mnot @ ( mbox @ c @ Phi ) )
                 @ ( mbox @ b @ ( mnot @ ( mbox @ c @ Phi ) ) ) ) ) )).

%----
thf(axiom_4,axiom,
    ( mvalid @ ( mnot @ ( mbox @ a @ ( ws @ a ) ) ) )).

thf(axiom_5,axiom,
```

```
        ( mvalid @ ( mnot @ ( mbox @ b @ ( ws @ b ) ) ) )).

%----
thf(conj,conjecture,
    ( mvalid @ ( mbox @ c @ ( ws @ c ) ) )).
```